\documentclass[conference]{IEEEtran}
\IEEEoverridecommandlockouts

\usepackage{cite}
\usepackage{amsmath,amssymb,amsfonts}
\usepackage{graphicx}
\usepackage{textcomp}
\usepackage{xspace}
\usepackage{xcolor}
\usepackage{booktabs} 
\usepackage{amsmath}
\usepackage{amsfonts}
\usepackage{amssymb}
\usepackage{amsthm}
\usepackage{comment}
\usepackage{footmisc}
\usepackage{mathrsfs}
\usepackage{graphicx}
\usepackage{subcaption}
\usepackage{mwe}
\usepackage{multirow}
\usepackage{algorithm}
\usepackage{algorithmicx}
\usepackage{algpseudocode}
\usepackage{multicol}
\usepackage{enumitem}
\usepackage{array}
\usepackage{float}
\usepackage{hyperref}
\usepackage{cleveref}
\usepackage{bbold}

\def\BibTeX{{\rm B\kern-.05em{\sc i\kern-.025em b}\kern-.08em
    T\kern-.1667em\lower.7ex\hbox{E}\kern-.125emX}}

\crefname{equation}{}{}
\Crefname{equation}{Equ}{Equations}

\newcommand{\best}[1]{$\bf{#1}^{\ast}$}
\newcommand{\vecb}[2]{\boldsymbol{#1}^{(#2)}}
\newcommand{\vect}[3]{\boldsymbol{#1}^{(#2)}_{#3}}
\newcommand{\subsubsect}[1]{\subsubsection{\textbf{#1}}}

\newcommand{\eg}{\emph{e.g.,}\xspace}
\newcommand{\ie}{\emph{i.e.,}\xspace}

\begin{document}

\newtheorem{ddf}{{\bf Definition}}
\newtheorem{thm}{{\bf Theorem}}

\title{A General Neural Causal Model for Interactive Recommendation}

\author{
\IEEEauthorblockN{Jialin Liu\textsuperscript{1,2}, Xinyan Su\textsuperscript{1,2}, Peng Zhou\textsuperscript{3}, Xiangyu Zhao\textsuperscript{4}, Jun Li\textsuperscript{1}}
\IEEEauthorblockA{
{\textsuperscript{1} Computer Network Information Center, Chinese Academy of Sciences, Beijing, China}\\
{\textsuperscript{2} University of Chinese Academy of Sciences, Beijing, China}\\
{\textsuperscript{2} Chinese University of Hong Kong, Hong Kong, China}\\
{\textsuperscript{4} City University of Hong Kong, Hong Kong, China}\\
{\{jlliu, suxinyan, lijun\}@cnic.cn, 1155023540@link.cuhk.edu.hk, xy.zhao@cityu.edu.hk}}
}

\maketitle

\begin{abstract}
Survivor bias in observational data leads the optimization of recommender systems towards local optima. Currently most solutions re-mines existing human-system collaboration patterns to maximize longer-term satisfaction by reinforcement learning. However, from the causal perspective, mitigating survivor effects requires answering a counterfactual problem, which is generally unidentifiable and inestimable. In this work, we propose a neural causal model to achieve counterfactual inference. Specifically, we first build a learnable structural causal model based on its available graphical representations which qualitatively characterizes the preference transitions. Mitigation of the survivor bias is achieved though counterfactual consistency. To identify the consistency, we use the Gumbel-max function as structural constrains. To estimate the consistency, we apply reinforcement optimizations, and use Gumbel-Softmax as a trade-off to get a differentiable function. Both theoretical and empirical studies demonstrate the effectiveness of our solution. 
\end{abstract}


\section{Introduction} \label{sec:introduction}

Recommendation systems accelerate many commercial applications by filtering out user-intended contents \cite{wan2018representing}. However, due to singularity of user preference, observed presentation only covers a fraction of the database,  with even less interaction recorded. This sparsity exacerbates the survivor bias \cite{lockwood2021fooled} when constructing effective recommendation policies. To further enhance the utility of recommendations, mitigating the survivor effect becomes essential.

In offline construction, historical behavior data becomes the survivor of observation, and the behavioral pattern can be sub-optimal since previously deployed systems used to collect the data is generally unknown \cite{chen2019top}. In online construction where limited experimental recommendation is allowed, the survivor effect still prevails because intervention can not be measured twice on the same user, whose state has been evolving since the first interaction. Under both offline and online scenarios, evaluating the survivor effect necessitates identifying a counterfactual question, \ie \textit{what if the system had previously chose another recommendation under the same state}. In real-world applications, counterfactual inference is challenging that acquires knowledge about the underline physical mechanism which we generally do not have \cite{pearl2018theoretical}.

Different types of user feedback reflects various aspects of their interests, \eg click signal reveals short-term preference during interaction, and purchase demonstrates a long-term preference usually coming after continuous clicks. To put both aspects into consideration, recent works \cite{chen2019generative,chen2021generative} frame the recommendation as Markov Decision Process (MDP) with recommender systems as agents and users as interactive environments to maximize long-term cumulation without sacrificing short-term utility\cite{bai2019model}. This Reinforcement Learning (RL) modulation can recombine high-value short-term transitions across different interaction trajectories to form higher long-term satisfaction \cite{rashidinejad2021bridging}, and thus alleviate survivor effects especially in offline environments. However, existing disciplines avoid directly answering the counterfactual question which is fundamental in offline RL research \cite{levine2020offline}.

In this work, we mitigate the survivor bias on the counterfactual hierarchy  \cite{bareinboim2022pearl}. Specifically, we first transform the measurement of different recommendation of the agent under current preference state into the consistency of same recommendation 
across different agents under the same preference, the latter can be formalized as the Probability of Necessity (PN) \cite{pearl2000causality} and benefit from the fact that we can use observational data to estimate the parametric agent when the PN is identifiable. The parameter space represents different agents. Causally, the survivor effect is reduced via counterfactual consistency. To identify the consistency with the ground true Structural Causal Model (SCM) unknown, we propose a general Neural Causal Model (NCM) based on the available graphical representation of the MDP, the proposed model uses learnable neural networks as an approximation of causal structural functions. Consistency is obtained via structural constraints, \ie Gumbel-max neural rewards. To estimate the proposed NCM, we implement a recursive neural architecture and three types of optimization procedures.

In a nutshell, our contributions are:
\begin{itemize}
    \item We propose a neural causal model to mitigate the survivor bias via consistency, the model is identifiable and estimable. Although we implement a vanilla neural architecture in this study, advanced regularization skills \cite{xiao2021general} can be bundled to further accelerate the performance.
    \item We theoretically prove the effectiveness of the proposed model. Empirical studies on both offline real-world datasets and online commercial simulators prove the generalization of this model. 
\end{itemize}
\section{Preliminaries}\label{sec:preliminaries}


\noindent \textbf{Notations}. $X$ denotes a random variable, $x$ represents its value. $X^{(t)}$ denotes timestamps and $Y_{k[x_k]}$ represents the $k$-th recommender system, \ie the factual system ($k=1$) and the counterfactual system ($k \neq 1$).

Graphically, SCM $\mathcal{M}$ is a Directed Acyclic Graph (DAG) of the tuple $<\boldsymbol{U}, \boldsymbol{V}, \mathcal{F}, P(\boldsymbol{U})>$. For $f_{V_i} \in \mathcal{F}$, a directed edge $(V_j \rightarrow V_i)$ maps $V_j \in \boldsymbol{Pa}_{V_i}$ to $V_i$. The corresponding neural SCM is defined as $\mathcal{M}(\boldsymbol{\theta}) \triangleq <\hat{\boldsymbol{U}}, \boldsymbol{V}, \hat{\mathcal{F}}, P(\hat{\boldsymbol{U}})>$, where structural functions $\mathcal{F}$ are parameterized with $\boldsymbol{\theta} = \{\boldsymbol{\theta}_{V_i}: V_i \in \boldsymbol{V}\}$,  each $\hat{f}_{V_i}$ is a FeedForward Neural Network (FFN). $\hat{\boldsymbol{U}}$ over bi-directed edges in $\mathcal{M}(\theta)$ is equivalently transformed into uniform prior according to neural causal theories \cite{xia2022neural}. With the language of $\mathcal{M}(\theta)$, counterfactual is formalized as a joint distribution over multiple interventions in $\mathcal{M}(\theta)$ \cite{berrevoets2021learning}.

\begin{ddf}[Counterfactual Distribution]
    \label{def:ctf-dist}
    Given $\mathcal{M}(\boldsymbol{\theta})$ and interventions
    $\boldsymbol{X} = \{\boldsymbol{X}_k: \boldsymbol{X}_k \subseteq \boldsymbol{V}, k = 1, \dots, K\}$,
    $P^{\mathcal{M}(\boldsymbol{\theta})}(\boldsymbol{Y}_{1[\boldsymbol{x}_1]}, \dots,
    \boldsymbol{Y}_{K[\boldsymbol{x}_K]})$ is computed as,
    \begin{equation*}
            \int_{\mathcal{D}_{\widehat{\boldsymbol{u}}}} \mathbb{1}\left[\boldsymbol{Y}_{1\left[\boldsymbol{x}_1\right]}(\widehat{\boldsymbol{u}})=\boldsymbol{y}_1,
            \dots, \boldsymbol{Y}_{K\left[\boldsymbol{x}_K\right]}(\widehat{\boldsymbol{u}})=\boldsymbol{y}_K\right]
            d P(\widehat{\boldsymbol{u}}),
    \end{equation*}
    where $\boldsymbol{Y}_{k\left[\boldsymbol{x}_k\right]}(\widehat{\boldsymbol{u}})$ comes
    from $\{f_{V_j}: V_j \in
    \boldsymbol{V} \setminus \boldsymbol{X}_k\} \bigcup \{f_X \leftarrow x: X \in
    \boldsymbol{X}_k\}$, and denotes the $k$-th imaginary interventions.
\end{ddf}

\textit{Probability of Necessity}\cite{pearl2000models} is formalized upon the counterfactual, which measures the necessity of current intervention to the observed results. In our task, the intervention is the recommendation policy and the result is the user feedback.
\begin{equation}
    \label{eq:pn}
    \textrm{PN}(\boldsymbol{Y}_{k\left[\boldsymbol{x}_k\right]} = \boldsymbol{y}) \triangleq
    P^{\mathcal{M}(\boldsymbol{\theta})}\bigl(\boldsymbol{Y}_{k\left[\boldsymbol{x}_k\right]} = \boldsymbol{y} \mid \boldsymbol{Y}_{1\left[\boldsymbol{x}_1\right]} = \boldsymbol{y}_1\bigr).
\end{equation}

We consider the recommendation task under the framework of MDP in this work, where users form the environment and recommender systems are the agents, recommendation is then an interaction between the agent and the environment, as the environment receives the action $\boldsymbol{a} \in \mathbb{R}^{|\mathcal{A}|}$ (recommendation), it adjusts its state $\boldsymbol{s} \in \mathbb{R}^{d_s}$ (user preference) and feeds rewards $r\left(\boldsymbol{s}, \boldsymbol{a}\right) \in \mathbb{R}$ (user behavior) back to the system $\pi_\theta(\boldsymbol{a} \mid \boldsymbol{s})$, which maximize discounted long-term satisfaction as:
\begin{equation} \label{eq:rl}
    \max _{\pi_\theta} \mathbb{E}_{\tau \sim \pi_\theta}\left[\sum_{t=0}^{|\tau|} \gamma^t r\left(\boldsymbol{s}^{(t)}, \boldsymbol{a}^{(t)}\right)\right],
\end{equation}
where $\tau=\left(\boldsymbol{s}^{(0)},\boldsymbol{a}^{(0)}, \dots, \boldsymbol{s}^{(|\tau|-1)}, \boldsymbol{a}^{(|\tau|-1)}\right)$ represents an episodic interaction. From causal perspective, the recommendation from the system to the user is an intervention from the agent to the environment, therefore different agents of the same environment represent different kinds of interventions. Even more challenging, different possible actions $\boldsymbol{a}_t$ of the same agents under same state $\boldsymbol{s}_t$ become counterfactual interventions ($k_1 \neq k_2$), because there can only be one observation and all other actions become counterfactual at the same time when the agent makes its recommendation choice.
\section{Framework}\label{sec:framework}
To mitigate survivor effect in both offline and online recommendation, we develop a general NCM in this section. We first introduce the model with theoretical analysis of its identifiability. Then we present a neural architecture which balances theoretical integrity and practical implementation. Finally, we present three common RL objective all of which can be used to optimize the neural architecture.

\begin{figure}[t]
    \centering
    \includegraphics[width=0.3\textwidth]{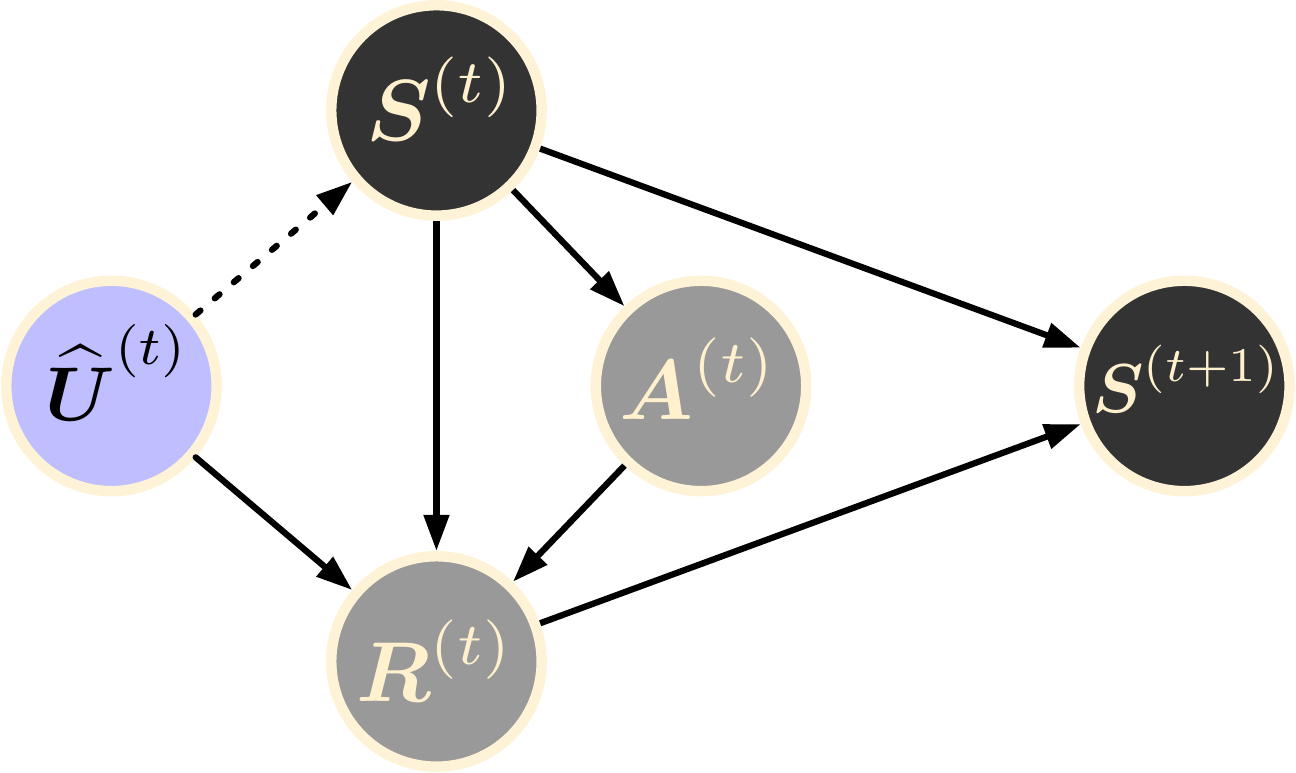}
    \caption{MDP as a neural-causal DAG. Blue cycles denote known exogenous $\widehat{\boldsymbol{U}} \sim \operatorname{Unif}[0,1]$, gray (observable) and black (computable thus observable) denotes endogenous $\boldsymbol{V} = \{\boldsymbol{S}, \boldsymbol{A}, \boldsymbol{R}\}$. Dash line $\widehat{\boldsymbol{U}} \dashrightarrow \boldsymbol{S}$ denotes a recursive relation due to Markov property useful to ease model implementation.}
    \label{fig:sess-cg}
\end{figure}

\subsection{Neural Causal Model}
As illustrated in \Cref{fig:sess-cg}, a $t-$time MDP can be untangled as a graphical representation of the NCM $\mathcal{M}(\theta_S, \theta_R, \theta_A)$,
\begin{equation} \label{eq:gm-ncm}
    \left\{\begin{array}{l}
        S^{(t)} = \hat{f}_S(S^{(t-1)}, R^{(t-1)}; \theta_S) \\
        R^{(t)} = \hat{f}_R(A^{(t)}, S^{(t)}, \widehat{U}^{(t)}; \theta_R) \\
        A^{(t)} = \hat{f}_A(S^{(t)}; \theta_A) \\
    \end{array}\right.,
\end{equation}
where structural function $\hat{f}$ is a parameterized unidirectional neural network. Considering the current state $\boldsymbol{s}^{(t)}$ of this causal model, different actions $\vect{a}{t}{k}$ ($k \neq 1 \in |\mathcal{A}|$) from the same agent $\pi_{1}$ is conceptually equivalent to the same action $\vect{a}{t}{f}$ from different agents $\pi_{k}$ which would all achieve the same user feedback $r_k(\vecb{s}{t}, \vect{a}{t}{1})$, since the policy is greedily searched \cite{chen2019generative}. The benefaction of this transformation is that $r_1(\vecb{s}{t}, \vect{a}{t}{1})$ is an instance of Survivor Effect \cite{lockwood2021fooled} and we can not observe other `survivors' $r_1(\vecb{s}{t}, \vect{a}{t}{k})$ especially in the offline environments. On-policy interaction (simulation) can alleviate this effect but this interaction is risky in recommendation \cite{schnabel2018short}. However, we can use a learnable agent $\pi_1$ in \cref{eq:gm-ncm} to approximate $r_k(\vecb{s}{t}, \vect{a}{t}{1})$ with consistency between $r_k$ and $r_1$ concerned. Formally, consistency can be described as the necessity of the current recommendation using \cref{eq:pn}:
\begin{equation}
    \label{eq:ncm-consistency}
    \begin{aligned}
    \frac{\textrm{PN}(R_{k[\vect{x}{t}{k}]} = r_1)}{P(R_{1[\vect{x}{t}{1}]} = r_1)} & \geqslant \frac{\textrm{PN}(R_{k[\vect{x}{t}{k}]} = r_k)}{P(R_{1[\vect{x}{t}{1}]} = r_k)} \\
    \Longrightarrow \textrm{PN}(R_{k[\vect{x}{t}{k}]} = r_k) & = 0,
    \end{aligned}
\end{equation}
where $\vect{x}{t}{k} = \{\boldsymbol{s},\vect{a}{t}{k}\}$ denotes the $k$-th option. This literally means to recommend current $\vect{a}{t}{1}$ under environment state $\vect{s}{t}{1}$, the reward $r_1$ needs to exceed other potentials. The key challenge is that \cref{eq:pn} is not directly estimable for statistic tools, \ie deep learning\cite{pearl2018theoretical, bareinboim2022pearl}, we need first identify its estimability, however, identification is generally not achievable \cite{pearl2000causality}. To be identifiable with consistency kept, we reduce model space of NCM $\mathcal{M}(\boldsymbol{\theta})$ with a particular reward form:
\begin{equation}
    \label{eq:gumbel-max-reward}
    R^{(t)} = \arg \max_r \left\{\log P_{\theta_R}\left(R^{(t)}=r \mid \vecb{s}{t}, \vecb{a}{t}\right) + g_r^{(t)} \right\},
\end{equation}
where $g_r^{(t)} = - \log (-\log (u_r^{(t)})$ and $u_r \sim \operatorname{Unif}(0,1)$. $P_{\theta_R} \varpropto exp\left(f_{\theta_R}(\boldsymbol{s}, \boldsymbol{a})\right)$ is a neural-estimated posterior. The restricted $\mathcal{M}'(\boldsymbol{\theta})$ now satisfies counterfactual-consistency \cref{eq:ncm-consistency}.

\begin{proof} \label{eq:proof}
According to identification theory \cite{xia2022neural}, $\mathcal{M}'(\boldsymbol{\theta})$ is neural identifiable if \cref{eq:ncm-consistency} is first symbolically identifiable and $\mathcal{M}'(\boldsymbol{\theta})$ matches the observational data $\mathbb{Z}_{data}$. To prove \cref{eq:gumbel-max-reward} satisfying the first requirement, suppose that $\forall r_1 \neq r_k \in \mathcal{D}_R$, then $R_{1[\boldsymbol{x}_1]} = r_1 \in \mathbb{Z}_{data}$ gives,
\begin{equation}
    \label{prof:r1}
    \log P(R_{1[\boldsymbol{x}]} = r_1) + g_{r_1} \geqslant \log P(R_{1[\boldsymbol{x}]} = r_k) + g_{r_k}.
\end{equation}
Now fix $r_k$ and make $P(R_{k[\boldsymbol{x}]} = r_k \mid R_{1[\boldsymbol{x}]} = r_1) \neq 0$ and get,
\begin{equation}
    \label{prof:rk}
    \begin{aligned}
    &\ \log P(R_{k[\boldsymbol{x}]} = r_k \mid R_{1[\boldsymbol{x}]} = r_1) + g_{r_k} \\
    \geqslant &\ \log P(R_{k[\boldsymbol{x}]} = r_1 \mid R_{1[\boldsymbol{x}]} = r_1) + g_{r_1}.
    \end{aligned}
\end{equation}
From \cref{prof:r1} and \cref{prof:rk}, we get this inequality,
\begin{equation*}
    \frac{\textrm{PN}(R_{k[a]} = r_1)}{P(R_{1[a]} = r_1)}
    \leqslant \frac{\textrm{PN}(R_{k[a]} = r_k)}{P(R_{1[a]} = r_k)},
\end{equation*}
the contrapositive proposition of \cref{eq:ncm-consistency}.
\end{proof}

We formalize a $\mathcal{M}'(\boldsymbol{\theta})$ with consistency concerned in this section, such formulation directly handles counterfactual queries which is bypassed before \cite{levine2020offline}. Theoretical analysis denotes $\mathcal{M}'(\boldsymbol{\theta})$ is identifiable. For the latter requirements in the proof, we develop an adversarial optimization whose convergence has been proved \cite{goodfellow2020generative}.

\subsection{Model Implementation}
 We address $\mathcal{M}'(\boldsymbol{\theta})$ implementation now. Specifically , we design major backbone in \cref{eq:gm-ncm} based on neural networks to benefit from the observational data collected.

\subsubsection{\textbf{Reward}} $\hat{f}_R$: Identifiability in \cref{eq:gumbel-max-reward} theoretically guarantees the estimablility of desired consistency \cref{eq:ncm-consistency}, yet this formulation is not differentiable. We replace $\arg \max$ with Gumbel-Softmax \cite{xiao2021general} as approximation which is:
\begin{equation}
    \label{eq:gumbel-softmax}
    \widetilde{R}^{(t)} \approx \frac{\exp \left(\left(\log \left(f_{\theta_R}\left(\vecb{s}{t}, \vecb{a}{t}\right)\right)+g_i\right) / \gamma_r\right)}{\sum_{j=1}^{|\mathcal{R}|} \exp \left(\left(\log \left(f_{\theta_R}\left(\vecb{s}{t}, \vecb{a}{t}\right)\right)+g_j\right) / \gamma_r\right)},
\end{equation}
where $f_\theta$ is a FFN due to neural-causal requirements, $\gamma_r$ is the balancing scalar, and $|\mathcal{R}|$ represents feedback types, \ie clicks, purchase and none. The trade-off in \cref{eq:gumbel-softmax} affects identifiability, however, it benefits model implementation and optimization since consistency itself serves as a kind of regularization which is adjustable in practice. We normalize \cref{eq:gumbel-softmax} to balance exploration and exploitation for policy learning:
\begin{equation}
    \label{eq:gsm-clipped-reward}
    r\left(\vecb{s}{t}, \vecb{a}{t};\theta_R\right) = 2 \times \widetilde{R}^{(t)} - 0.5.
\end{equation}
Based on \cref{eq:gsm-clipped-reward}, we now tackle the matching issue required with adversarial estimation. Specifically, \Cref{eq:gsm-clipped-reward} is implemented as a discriminator $D(r^{(t)}_{k[\boldsymbol{a}_k]} \mid \vecb{s}{t}, \vecb{a}{t};\theta_D)$. As consistency \cref{eq:ncm-consistency} guarantees $r^{(t)}_{k[a]}=r^{(t)}_{1[a]}$ at dynamic equilibrium, the overall optimization is defined as:
\begin{equation}
    \label{eq:al}
    \begin{aligned}
    & \min_{\pi_1} \max_D \mathbb{E}_{r^{(t)} \sim p\left(R_{1[a_1]}\right)}\left[\log D\left(r_{1[a_1]}^{(t)} \mid \vecb{s}{t}, \vecb{a}{t};\theta_D\right)\right] \\
    + &\ \mathbb{E}_{\hat{u}^{(t)} \sim P\left(\widehat{U}\right)}\left[\log \left(1 - D\left(r_{1[a_1]}^{(t)} \mid \vecb{s}{t}, \vecb{a}{t};\theta_D\right)\right)\right],
    \end{aligned}
\end{equation}
we drop the state $\vect{s}{t}{1}$ in $\vect{x}{t}{1}$ for simplicity.

\subsubsection{\textbf{Agent}} $\hat{f}_A$: The recommender agent aims to generate relevant candidates for user to browse. Since available options are generally huge in modern recommendation platforms ($|\mathcal{A}| \gg 1$), we implement the factual recommending agent upon $\hat{f}_S$ to ease model complexity:
\begin{equation}
    \vect{h}{t}{1[\boldsymbol{a}_i]} = \boldsymbol{w}_{(a)}^T\sigma\left(\boldsymbol{W}_{(a)}\left[\left(\vecb{s}{t}\right)^T, \left(\vect{a}{t}{i}\right)^T\right]^T + \boldsymbol{b}_{(a)}\right),
\end{equation}
where $\theta_A = \{\boldsymbol{w}_{(a)}, \boldsymbol{W}_{(a)}, \boldsymbol{b}_{(a)}\}$. As \Cref{fig:sess-cg} denotes, the recursion comes from the fact that \cref{eq:gm-ncm} is bundled upon each other. Consequently, we implement the agent as follows:
\begin{equation}
    \label{eq:agent}
    \begin{aligned}
    \pi\left(i \in \vecb{a}{t} \mid \vecb{s}{t}; \theta_A\right)
    = \frac{\exp \left(\left(\log \vect{h}{t}{1[\boldsymbol{a}_i]} +g_i\right) / \gamma_a\right)}{\sum_{j=1}^{|\mathcal{A}|} \exp \left(\left(\log \vect{h}{t}{1[\boldsymbol{a}_j]} + g_j\right) / \gamma_a\right)},
    \end{aligned}
\end{equation}
where $\{g_j\}_{j=1}^{|\mathcal{A}|}$ is i.i.d. sampled from Gumbel distribution with scalar $\gamma_a$, here we use Gumbel-softmax again.

\subsubsection{\textbf{State}} $\hat{f}_S$: This function aims to encode preference transitions in \cref{eq:gm-ncm} according to previous browsing history. Attention mechanism has been proved effective to capturing this autoregressive dynamics \cite{kang2018self}, thus we apply here for $\hat{f}_S$ estimation. Position codings  are involved since self-attention does not contain temporal information. Specifically, $\hat{f}_S$ first encodes hitherto interactions $\boldsymbol{i} := [r_{1[a]}^{(0)}, r_{1[a]}^{(1)}, \dots, r_{1[a]}^{(t-1)}]$ as the matrix $\boldsymbol{E}$. Then,position-aware matrix $\boldsymbol{P}$ is learned as:
\begin{equation*}
  \widehat{\boldsymbol{E}} = \boldsymbol{E} + \boldsymbol{P},
\end{equation*}
where $\boldsymbol{E} \in \mathbb{R}^{n \times d}$ and $\boldsymbol{P} \in \mathbb{R}^{n \times d}$ of $d$ dimension. Based on it, the dot-product attention \cite{vaswani2017attention} computes a weighted sum scaled by the dimensional factor,
\begin{equation*}
  \operatorname{Att}(\boldsymbol{Q}, \boldsymbol{K}, \boldsymbol{V}) = \operatorname{Softmax}\left(\boldsymbol{Q K}^T / \sqrt{d}\right) \boldsymbol{V},
\end{equation*}
where $\boldsymbol{Q}, \boldsymbol{K}, \boldsymbol{V}$ denotes query, key and value matrices. For our task, each of three matrices is linearly projected from the same $\widehat{\boldsymbol{E}}$, a multi-head layer is then concatenated as,
\begin{equation*}
  \begin{aligned}
    & \boldsymbol{H} = \operatorname{SA}(\boldsymbol{\widehat{E}}) = \left[\text{head}_1 ; \cdots ; \text{head}_h\right] \boldsymbol{W}, \\
    & \operatorname{head}_\ell = \operatorname{Att}\left(\boldsymbol{\widehat{E} W}_\ell^Q, \boldsymbol{\widehat{E} W}_\ell^K, \boldsymbol{\widehat{E} W}_\ell^V\right),
    \end{aligned}
\end{equation*}
where $\boldsymbol{W}_\ell^Q, \boldsymbol{W}_\ell^K, \boldsymbol{W}_\ell^V \in \mathbb{R}^{d \times \frac{d}{h}}$ and $\boldsymbol{W} \in \mathbb{R}^{d \times d}$ are are weights of the $\ell-$th head. To avoid future leaking which is anti-causal, we mask out links between $\boldsymbol{Q}_i$ and $\boldsymbol{K}_j$ where $j > i$, and encourage asymmetry via a point-wise FFN:
\begin{equation*}
  \boldsymbol{S}_i = \operatorname{FFN}(\boldsymbol{H}_i) = \operatorname{ReLU}(\boldsymbol{H}_i\boldsymbol{W}^{(f_1)} + \boldsymbol{b}^{(f_1)})\boldsymbol{W}^{(f_2)} + \boldsymbol{b}^{(f_2)},
\end{equation*}
where $\boldsymbol{W}^{(f_1)}, \boldsymbol{W}^{(f_2)} \in \mathbb{R}^{d \times d}$ and $\boldsymbol{b}^{(f_1)}, \boldsymbol{b}^{(f_2)} \in \mathbb{R}^{d}$. The $b$-th self-attention block is designed as,
\begin{equation} \label{eq:multi-head-attention-blocks}
  \begin{aligned}
    & \boldsymbol{H}^{(b)} = \operatorname{SA}(\boldsymbol{S}^{(b-1)}), \\
    & \boldsymbol{S}_j^{(b)} = \operatorname{FFN}(\boldsymbol{H}^{(b)}_j),
  \end{aligned}
\end{equation}
where $j \in \{1, 2, \dots, n\}$ denotes the first $j$ items. Note that we reorganize reward feedback (\ie pass, click and purchase) into binary groups (\ie pass or not) to ease implementation, extension to involve mult-type valuation can be achieved via regression upon self-attention blocks.

\begin{algorithm}[t]
    \caption{\label{alg} Model Optimization.}
	\raggedright
	{\bf Input}: Observation $\mathcal{O} = \{[r^{(0)}_{1[x]}, \dots, r^{(T_m)}_{1[x]}]\}_{m=1}^{M}$.\\
	\begin{algorithmic} [1]
        \State Initialize parameters $\theta_D, \theta_A, \theta_S, \theta_V$.
        \For{iteration $i = 0, 1, \dots$}
            \For {step $j = 0, 1, \dots$}
                \State Sample exogenous priors $u_r \sim P(\widehat{U})$.
                \State Sample observational trajectories $(\boldsymbol{s}, \boldsymbol{a}) \sim \mathcal{O}$
                \State Update discriminator parameters $\theta_D, \theta_S$  \Comment{\cref{eq:al}}
            \EndFor
            \For {step $k = 0, 1, \dots$}
                \State Sample interventional trajectories $(\boldsymbol{s}, \boldsymbol{a}) \sim \pi_{\theta_A}$
                \State Sample exogenous priors $u_r \sim P(\widehat{U})$.
                \State Update parameters $\theta_A, \theta_V$  \Comment{\cref{eq:pl,eq:td,eq:ac}}
            \EndFor
        \EndFor
	\end{algorithmic}
\end{algorithm}

\subsection{Model Optimization}
$\mathcal{M}'(\boldsymbol{\theta}) \in \mathcal{M}(\boldsymbol{\theta})$ is still an expressive model which can benefit common reinforcement optimizations, \eg policy-based learning \cite{bai2019model}, value-based \cite{chen2019generative} and actor-critic learning \cite{chen2021generative}. We consider these representatives in our studies as a demonstration of generality of $\mathcal{M}'(\boldsymbol{\theta})$. Specifically, for both optimizations, we use \cref{eq:agent} as the policy network and a FFN as the critic network. \Cref{alg} shows the overall optimization.

\subsubsect{Policy-based Learning} we adopt policy REINFORCE \cite{bai2019model} as an illustration. This learning approach directly optimize \cref{eq:rl} with reparameterization tricks with following gradients:
\begin{equation}
    \label{eq:pl}
    \mathbb{E}_{\tau \sim data} \left[\sum_{t=0}^{|\tau|} V^{(t)} \nabla_{\theta_A} \log \pi\left(i \in \vecb{a}{t} \mid \vecb{s}{t}; \theta_A\right)\right],
\end{equation}
where $V^{(t)} = \sum_{t'=t}^{t'=|\tau|} \gamma^{t' - t} D(\vecb{s}{t}, \vecb{a}{t})$ is the critic valuation. \cref{eq:pl} is originally designed for online interaction, since \cref{eq:rl} requires on-policy evaluations. In offline environments, off-correction \cite{chen2019top} can alleviate distribution shifting.

\subsubsect{Value-based Learning} we utilize Temporal Difference (TD) \cite{chen2019generative} as a representative, which updates the critic network $V(\vecb{s}{t}, \vecb{a}{t})$ with following gradients:
\begin{equation}
    \label{eq:td}
    \begin{aligned}
        \mathbb{E}_{\tau \sim data} & \left[\nabla_{\theta_V}\left(D(\vecb{s}{t}, \vecb{a}{t}) + \gamma \max _{a^{\prime}} V\left(\vecb{s}{t+1}, \boldsymbol{a}^{\prime} ;\theta_V \right) \right.\right. \\
        - & \left.\left.V\left(\vecb{s}{t}, \vecb{a}{t} ;\theta_V\right)\right)^2\right],
    \end{aligned}
\end{equation}
where $\theta_V$ contains policy parameters $\theta_A$ and the parameters from the critic FNN $V(\boldsymbol{s}, \boldsymbol{a})$ itself.

\subsubsect{Actor-Critic Learning} we consider General Advantage Estimation (GAE) \cite{chen2021generative} to update a proximal policy objective:
\begin{equation}
    \label{eq:ac}
    \mathbb{E}_{\tau \sim data} \left[\min \left(\frac{\pi_\theta}{\pi_{\theta_{\text{old }}}} A^{(t)}, \operatorname{clip}\left(\frac{\pi_\theta}{\pi_{\theta_{\text{old }}}}, 1-\epsilon, 1+\epsilon\right) A^{(t)}\right)\right],
\end{equation}
where $\epsilon$ is the clipping scalar for the conservative updates. The cumulative advantage function $A$ is estimated as follows:
\begin{equation}
    \label{eq:advantage}
    A^{(t)} = \sum_{l=0}^{\infty}\left(\gamma \lambda_g\right)^l\left[-V\left(\vecb{s}{t}\right)+\sum_{l=0}^{\infty} \gamma^l D^{(t+l)}\right],
\end{equation}
where $l$ denotes the $l$-step away from the current time $t$.

As a summary, we formalize a counterfactual-consistent NCM in this section to mitigate the survivor effect, and propose to use recursion and softmax trade-off for implementation. We also theoretically prove the consistency is identifiable. Empirical studies in next section also demonstrate the effectiveness of these designs.
\section{experiments}
To verify the mitigation on survivor effect, we conduct experiments with two consideration: (i) \textbf{Generalization}. Does the counterfactual consistency is effective in both offline learning and online learning? (ii) \textbf{Adaptivity}. Can different optimization procedures all benefit the mitigation effect?

\subsection{Experimental Setup}
\noindent \textbf{Data.} Offline experiments are conducted on two released recommendation datasets \ie \textit{Kaggle}\footnote{\url{https://www.kaggle.com/retailrocket/ecommerce-dataset}} and \textit{RecSys15}\footnote{\url{https://recsys.acm.org/recsys15/challenge}}. For online experiments, we use a  simulator \textit{VirtualTB\footnote{\url{https://github.com/eyounx/VirtualTaobao.git}}}.
\begin{itemize}
    \item \textbf{Offline datasets} We treat views as clicks and adding items to the cart as purchases resulting in binary user feedbacks, \ie clicks and purchases. Both items and interaction trajectories less than 3 times is removed because of sparsity. \Cref{table:data} details the preprocessing results.
    \item \textbf{Online simulations} \textit{VirtualTB} simulates real-world user behaviors on one of the largest online e-commerce platforms. In this simulator, each user has 11 binary attributes encoded as an 88-dim vector, and recommendation action is a 27-dimensional vector, the immediate reward as feedback signals are integers from 0 to 10.
\end{itemize}

\noindent \textbf{Metrics.} For offline evaluation, we measure top-k $(k=\{5, 10\})$ Hit Ratio (H@k) \cite{xin2020self} and Normalized Discounted Cumulative Gain (N@k) \cite{jarvelin2002cumulated}, widely adopted as a measurement for recalling and ranking \cite{xiao2021general,xin2020self}. For online simulations, we use click-through-rate from the simulator as:
\begin{equation*}
    CTR = \frac{r_{epi}}{10 \times N_{epi}},
\end{equation*}
where $r_{epi}$ is the episodic rewards in $N_{epi}-$length interactions.

\noindent \textbf{Baselines.} We consider three types of learning strategies: IRe \cite{bai2019model}, CQN \cite{chen2019generative} and Inv \cite{chen2021generative}. These methods all adopt adversarial learning combined with reinforcement learning to optimize reward functions, IRe inducts REINFORCE (policy-based) to learn the recommendation agent. CQN applies Temporal Difference (value-based) to implicitly update the agent. InvGAN uses Generative Advantage Estimation (actor-critic based) to update state functions. The major difference is the reward function formulation, where IRe and CQN utilize FNN, while Inv sets the log-scale discrimination difference as the reward that can involve additive reward for enhancement. 

\noindent \textbf{Implementation.} All bases use the same neural architecture. Since offline datasets offer expertise demonstration and we use a unified policy collector in online simulations, we implement a model-free IRe and CQN for simplicity. For CQN, we use supervised regularization, \ie cross-entropy, in addition with the adversarial learning \cref{eq:al}. For Inv, we add predefined feedback in addition with the differential reward, \ie 0.2 for click and 1.0 for purchase, 0.0 otherwise, this setting is proved effective \cite{xin2020self}. For online simulators, we adopt Deep Deterministic Policy Gradients (DDPG) as original IncRec suggests to train the expert policy collector. One-head self-attention block with embedding size 50 is adopted, the learning rate for the actor is $1e-4$ and $1e-3$ for the critic, both of which are optimized with Adam \cite{kingma2014adam}. The discounted factor $\gamma$ is 0.7. We use 10 recent interactions as input length ($w=10$), with mini-batch $B=256$. Item embeddings are initialized from Gaussian distribution. For the agent \cref{eq:agent}, we adopt a 2-layer FNN with with 512 hidden units and ReLU as nonlinear activation, $\gamma = 0.2$ for Gumbel-Softmax, $\lambda_g = 0.97$ and $\epsilon = 0.2$ for GAE \cite{chen2021generative}, same FNN for the critic $V$. 100 episodes as 1 iteration in VirtualTB for illustration.

\begin{table}[t]
  \centering
  \caption{Data Statistics.}
  \setlength{\tabcolsep}{30pt}
  \label{table:data}
  \scalebox{1.0}{
    \begin{tabular}{@{}lll@{}}
      \toprule[1pt]
      & \textit{Kaggle} & \textit{RecSys15} \\
      \midrule
      \#interactions & 195,523 & 200,000 \\
      \#items & 70,852 & 26,702 \\
      \#clicks & 1,176,680 & 1,110,965 \\
      \#purchases & 57,269 & 43,946 \\
      \bottomrule[1pt]
    \end{tabular}
  }
  \vspace{-4mm}
\end{table}

\subsection{Experimental Results}
\subsubsect{Offline Performance} \Cref{table:click/att} and \Cref{table:purchase/att} details offline performance. First, we observe that Inv works worst, because GAE used by Inv in its nature is an on-policy evaluation which acquires distribution correction to use trajectories collected by other policies, and TD used by CQN offers an off-policy evaluation more suitable for offline environments. Second, we observe each type of reinforcement optimization methods can benefit from the neural causal model, specifically, the Gumbel reward implementation, since offline environment can not cover each possible state-action-reward tuples and online interaction is unavailable. Technically, Gumbel design here serves as regularization to reduce model complexity. Theoretically, the regularization comes from consistency between current policy (factual) and potential policies (counterfactual), this consistency as a causal quantity is identifiable and thus can be estimated without conducting random control experiments, which is necessary for unidentifiable situations. In recommendation task, random control experiments are unrestricted online interaction which is unrealistic for safety concerned \cite{schnabel2018short}.

\begin{table}[t]
  \centering
  \caption{Offline performance. Bold denotes the best. ``{\bf $\ast$}'' denotes the statistically significant improvements (\ie two-sided t-test with $p<0.05$) over the best baseline.}
  \label{table:click/att}
  \scalebox{.77}{
    \begin{tabular}{@{}lllllllll@{}}
      \toprule[1pt]
      \multirow{ 2}{*}{click} & \multicolumn{4}{c}{RecSys} & \multicolumn{4}{c}{Kaggle} \\ \cmidrule(lr){2-5} \cmidrule(lr){6-9}
      & H@5 & N@5 & H@10 & N@10 & H@5 & N@5 & H@10 & N@10 \\
      \midrule
        Inv   & .3159 & .2179 & .4142	& .2503 & .2700 & .2119 & .3238 & .2236 \\
        Inv+  & \best{.3269} & \best{.2312} & \best{.4360} & \best{.2612} & \best{.2760} & \best{.2136} & \best{.3291} & \best{.2281} \\
        \midrule
        IRe   & .3339 & .2318 & .4374	& .2672 & .2775 & .2163 & .3394 & .2319 \\
        IRe+  & \best{.3361} & \best{.2385} & \best{.4438}	& \best{.2686} & \best{.2882} & \best{.2202} & \best{.3564} & \best{.2408} \\
        \midrule
        CQN   & .3343 & .2364 & .4451	& .2850 & .2912 & .2249 & .3512 & .2369 \\
        CQN+  & \best{.3571} & \best{.2473} & \best{.4619}	& \best{.3154} & \best{.3029} & \best{.2458} & \best{.3677} & \best{.2614} \\
      \bottomrule[1pt]
    \end{tabular}
  }
\end{table}

\begin{table}[t]
  \centering
  \caption{Offline performance. Bold denotes the best. ``{\bf $\ast$}'' denotes the statistically significant improvements (\ie two-sided t-test with $p<0.05$) over the best baseline.}
  \label{table:purchase/att}
  \scalebox{.77}{
    \begin{tabular}{@{}lllllllll@{}}
      \toprule[1pt]
      \multirow{ 2}{*}{purchase} & \multicolumn{4}{c}{RecSys} & \multicolumn{4}{c}{Kaggle} \\ \cmidrule(lr){2-5} \cmidrule(lr){6-9}
      & H@5 & N@5 & H@10 & N@10 & H@5 & N@5 & H@10 & N@10 \\
      \midrule
        Inv	& .4051 & .2605	& .5438	& .3296	& .5233	& .4397	& .5985	& .4542 \\
        Inv+	& \best{.4252} & \best{.2750}	& \best{.5603}	& \best{.3693}	& \best{.5382}	& \best{.4569}	& \best{.6252}	& \best{.4850} \\
        \midrule
        IRe	& .4209 & .3039	& .5569	& .3516	& .5306	& .4406	& .6264	& .4748 \\
        IRe+	& \best{.4416} & \best{.3104}	& \best{.5763}	& \best{.3738}	& \best{.5383}	& \best{.4592}	& \best{.6392}	& \best{.4929} \\
        \midrule
        CQN	& .4327 & .3082	& .5603	& .3791	& .5327	& .4572	& .6272	& .4787 \\
        CQN+	& \best{.4494} & \best{.3213}	& \best{.5779}	& \best{.3941}	& \best{.5483}	& \best{.4674}	& \best{.6433}	& \best{.4961} \\
      \bottomrule[1pt]
    \end{tabular}
  }
  \vspace{-4mm}
\end{table}

\begin{figure}[htbp]
    \begin{subfigure}{.45\linewidth}
        \centering
        \includegraphics[height=.17\textheight]{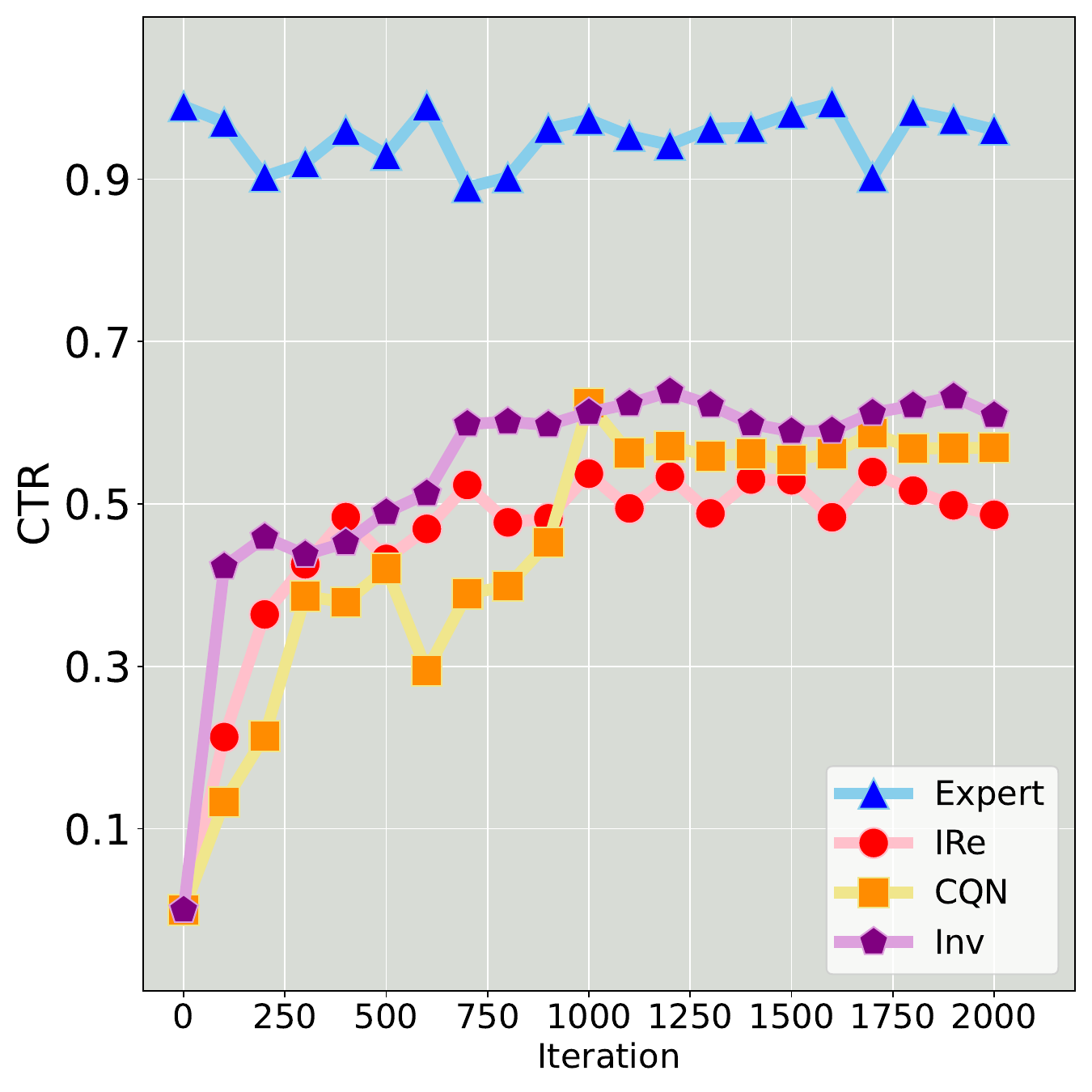}
        \caption{}
        \label{fig:base}
    \end{subfigure}
    \begin{subfigure}{.45\linewidth}
        \centering
        \includegraphics[height=.17\textheight]{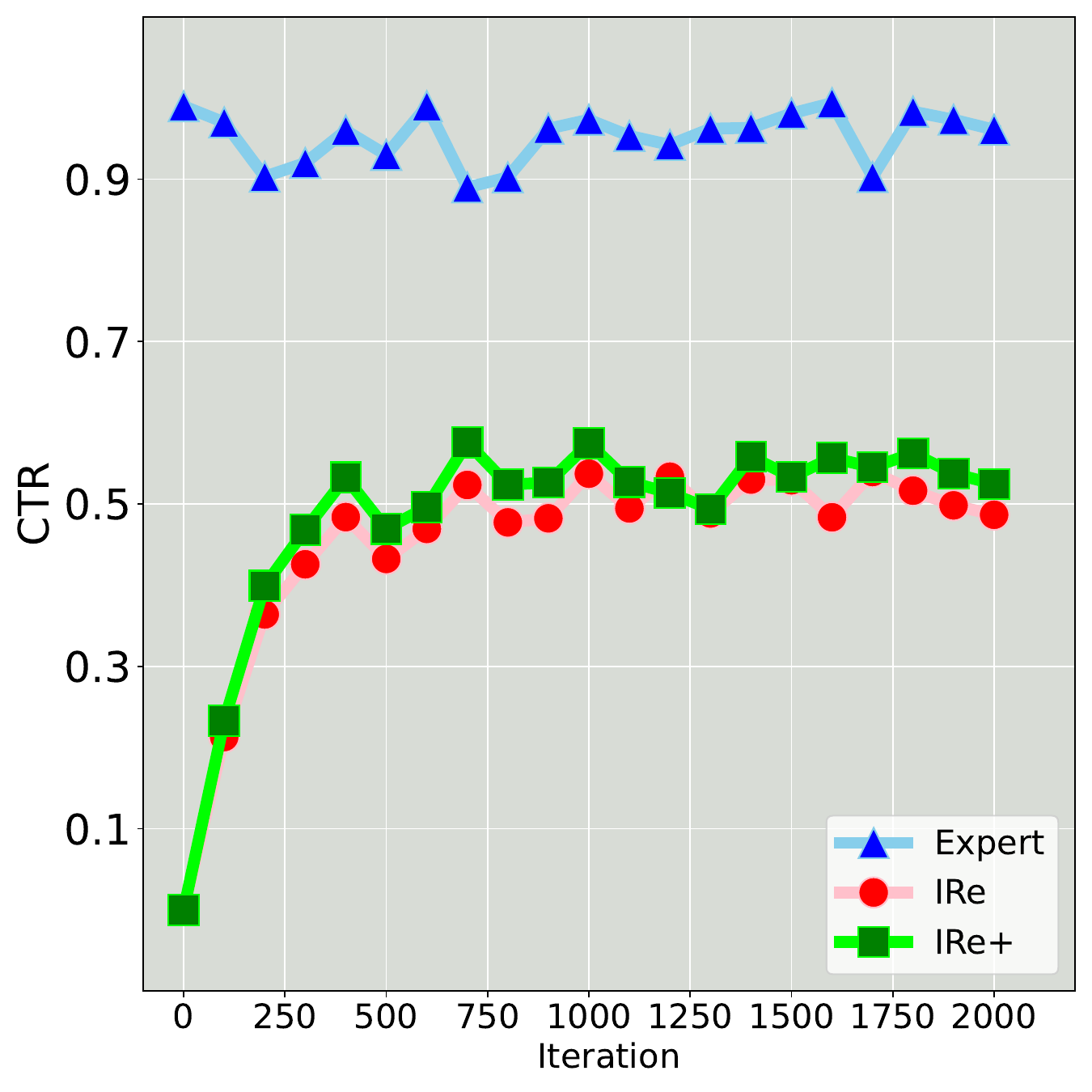}
        \caption{}
        \label{fig:irecgan}
    \end{subfigure}
    \\
    \begin{subfigure}{.45\linewidth}
        \centering
        \includegraphics[height=.17\textheight]{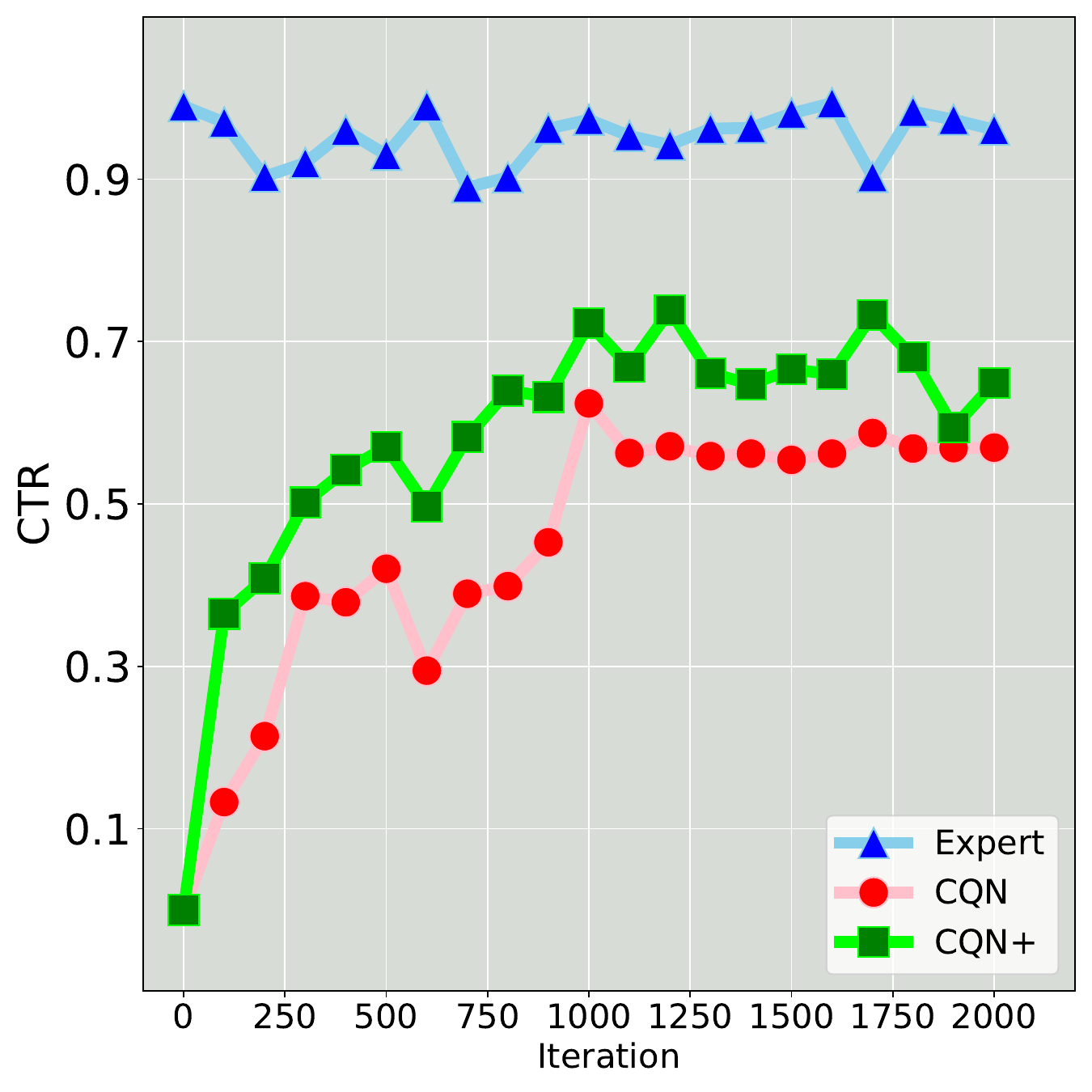}
        \caption{}
        \label{fig:cascadedqn}
    \end{subfigure}
    \begin{subfigure}{.45\linewidth}
        \centering
        \includegraphics[height=.17\textheight]{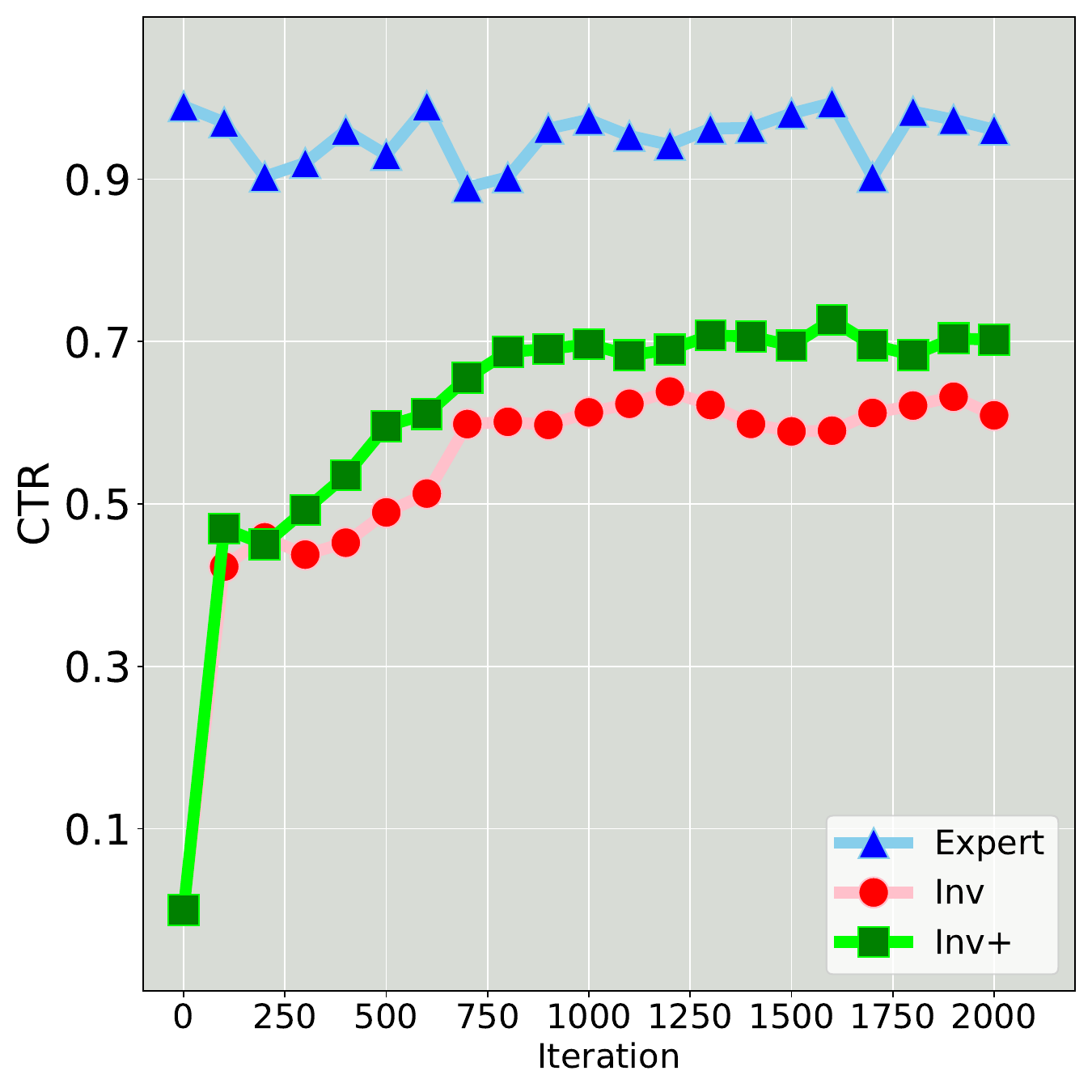}
        \caption{}
        \label{fig:invrec}
    \end{subfigure}
    \caption{Online simulations with VirtualTB.}
    \label{fig:online}
    \vspace{-4mm}
\end{figure}
\subsubsect{Online Performance} Simulator offers an environment to timely interact. \Cref{fig:online} details simulation results, where blue lines ('Expert') is the DDPG expert collectors. \Cref{fig:base} compares among baselines, we observe that Inv works best different from offline experiments, since it can immediately evaluate the agent with the simulator without further restriction, and the reward is learnt by imitating expert policy from DDPG. All three baselines are improved with Gumbel regularization, which proves that regularization via consistency is still effective with free interaction available, because counterfactual consistency is  inheritable via lower interventional level (online) and observational level (offline)\cite{bareinboim2022pearl} moreover, this also demonstrates Gumbel-Softmax trade-off is still effective while it hurts theoretical integrity. We also observe more imporvement over CQN and Inv than IRe, this benefaction comes from the implicit consistency on cumulative valuation, \ie Q-value in CQN and advantage function in Inv.

As a brief summary, empirical results verify $\mathcal{M}'(\theta)$ is \textbf{generalizable} and \textbf{adaptive} , since all three baselines are improved in both offline and online experiments. Furthermore, the recursion we use to reduce model complexity and the softmax trade-off is proved effective.

\section{Related Works}

Collaborative recommendation \cite{rendle2010factorizing} fails to capture high-order interaction relations. Autoregressive approaches \cite{kang2018self} model this relation as evolving sequences. For multiple feedbacks, reinforcement recommenders maximize cumulative valuation as the representative of user satisfaction, existing works covers policy-based methods with distribution correction \cite{chen2019top,bai2019model}, value-based methods\cite{xin2020self,chen2019generative} with supervised or adversarial regularization, and actor-critic methods \cite{xiao2021general} with action space approximation. Typically, reward functions is predefined. To automate the reward fine-tuning procedure, recent works learn to reward inversely from user feedbacks \cite{chen2021generative}. Both reinforcement agents need correction since offline collections can be inconsistent with online interaction \cite{schnabel2018short}. In this work, we develop a counterfactual-consistent causal model based on recent development of causal inference. Causality formalize the language to analyze model inductive bias \cite{pearl2018theoretical, bareinboim2022pearl}, and derive desired properties, \ie consistency in this work, as a mathematical quantity. Recent research develope causal machine learning methods, \ie neural-causal connection \cite{xia2021causal}, which is the foundation of our method.
\section{conclusion}
In this work, we propose a novel NCM which theoretically achieves consistency on counterfactual hierarchy. Such consistency is necessary to tackle survivor effect in recommendation. We implement our model with major RL methods. Empirical experiments on both offline and online environments prove the adaptivity and generalization. We assume observable MDP in this studies, partially-observed MDP will be explored next.
\section*{acknowledgement}
This research is supported by APRC - CityU New Research Initiatives (No.9610565, No.9360163), Hong Kong ITC Fund Project (No.ITS/034/22MS), and SIRG - CityU Strategic Research Grant (No.7020046, No.7020074, No.7005894).

\bibliographystyle{IEEEtran}
\bibliography{IEEEabrv, 7_simplified_references}

\end{document}